%% file: CVPR17_Bosch.tex
\documentclass[10pt,twocolumn,letterpaper]{article}

\usepackage{iccv}
\usepackage{times}
\usepackage{epsfig}
\usepackage{graphicx}
\usepackage{amsmath}
\usepackage{amssymb}
\usepackage{multirow}
\usepackage{tabularx}
\usepackage{algpseudocode}
\usepackage[ruled,vlined,linesnumbered]{algorithm2e}
\usepackage{breqn}
\usepackage{soul,xcolor}
\usepackage{cite}
\usepackage{authblk}

\def\figvspace{{\vspace{-3mm}}}
\newcommand{\Paragraph}[1]{\vspace{-0mm} \noindent \textbf{#1} \hspace{0mm}}
\newcommand{\Section}[1]{\vspace{-1mm} \section{#1} \vspace{-0mm}}
\newcommand{\SubSection}[1]{\vspace{-1mm} \subsection{#1} \vspace{-0mm}}

\usepackage[pagebackref=true,breaklinks=true,letterpaper=true,colorlinks,bookmarks=false]{hyperref}

 \iccvfinalcopy 

\def\iccvPaperID{1488} 

\ificcvfinal\fi
\begin{document}


\title{Pose-Invariant Face Alignment with a Single CNN}

\author[1]{Amin Jourabloo}
\author[2]{Mao Ye}
\author[1]{Xiaoming Liu}
\author[2]{Liu Ren}
\affil[1]{Department of Computer Science and Engineering, Michigan State University}
\affil[2]{Visualization Group, Bosch Research and Technology Center North America}
\affil[1,2]{ {\{jourablo, liuxm\}@msu.edu}, {\{mao.ye2, liu.ren\}@us.bosch.com}}


\maketitle


\begin{abstract}
Face alignment has witnessed substantial progress in the last decade. 
One of the recent focuses has been aligning a dense $3$D face shape to face images with large head poses. 
The dominant technology used is based on the cascade of regressors, e.g., CNN, which has shown promising results. 
Nonetheless, the cascade of CNNs suffers from several drawbacks, e.g., lack of end-to-end training, hand-crafted features and slow training speed. 
To address these issues, we propose a new layer, named visualization layer, that can be integrated into the CNN architecture and enables joint optimization with different loss functions. 
Extensive evaluation of the proposed method on multiple datasets demonstrates state-of-the-art accuracy, while reducing the training time by more than half compared to the typical cascade of CNNs. 
In addition, we compare multiple CNN architectures with the visualization layer to further demonstrate the advantage of its utilization. 
\end{abstract}

\vspace{-5mm}

\input{CVPR17_Bosch_intro.tex}
\input{CVPR17_Bosch_prior.tex}

\input{CVPR17_Bosch_proposed.tex}

\input{CVPR17_Bosch_exp.tex}
\input{CVPR17_Bosch_con.tex}

\clearpage

{\small
\bibliographystyle{ieee}
\bibliography{abbrev_brief,egbib}
}

\end{document}

%% file: CVPR17_Bosch_intro.tex
\Section{Introduction}
\label{Sec:Intro}

Face alignment, also known as face landmark detection, is an essential process for many facial analysis tasks, such as face recognition~\cite{wagner2012toward}, expression estimation~\cite{bettadapura2012face} and $3$D face reconstruction~\cite{kemelmacher2011face,roth2016adaptive}. 
During the last decade, face alignment technologies have been substantially improved~\cite{cootes2001activea,saragih2009facea,cristinacce2007boosteda,xiong2013supervised,cao2014face}. 
One recent advancement in this area is to tackle challenging cases with large face poses, e.g., frontal to profile views with $\pm90^{\circ}$ yaw angles~\cite{jourabloo2016large,jourabloo2017pose,zhu2015face,liu2016joint,zhao2016fast,mcdonagh2016joint}. 


The dominant technology for large-pose face alignment (LPFA) utilizes a cascade of regressors which combines different types of regression designs~\cite{liu2016joint,zhuunconstrained,zhu2015face} with feature extraction methods~\cite{jourabloo2016large}. 
At each stage of this procedure, the target parameters, e.g., $2$D landmarks or the head pose and $3$D face shape, are refined by regressing an update of these parameters. 
Due to the proven power of Convolutional Neural Network (CNN) in vision tasks, it is also adopted as the regressor in this framework and has achieved the state-of-the-art performance on face alignment~\cite{jourabloo2016large,zhu2015face,trigeorgis2016mnemonic,liu2016joint}. 

\begin{figure}[t!]\small
\begin{center}
\begin{tabular}{ccccc}
\multicolumn{5}{c}{\includegraphics[width=\linewidth]{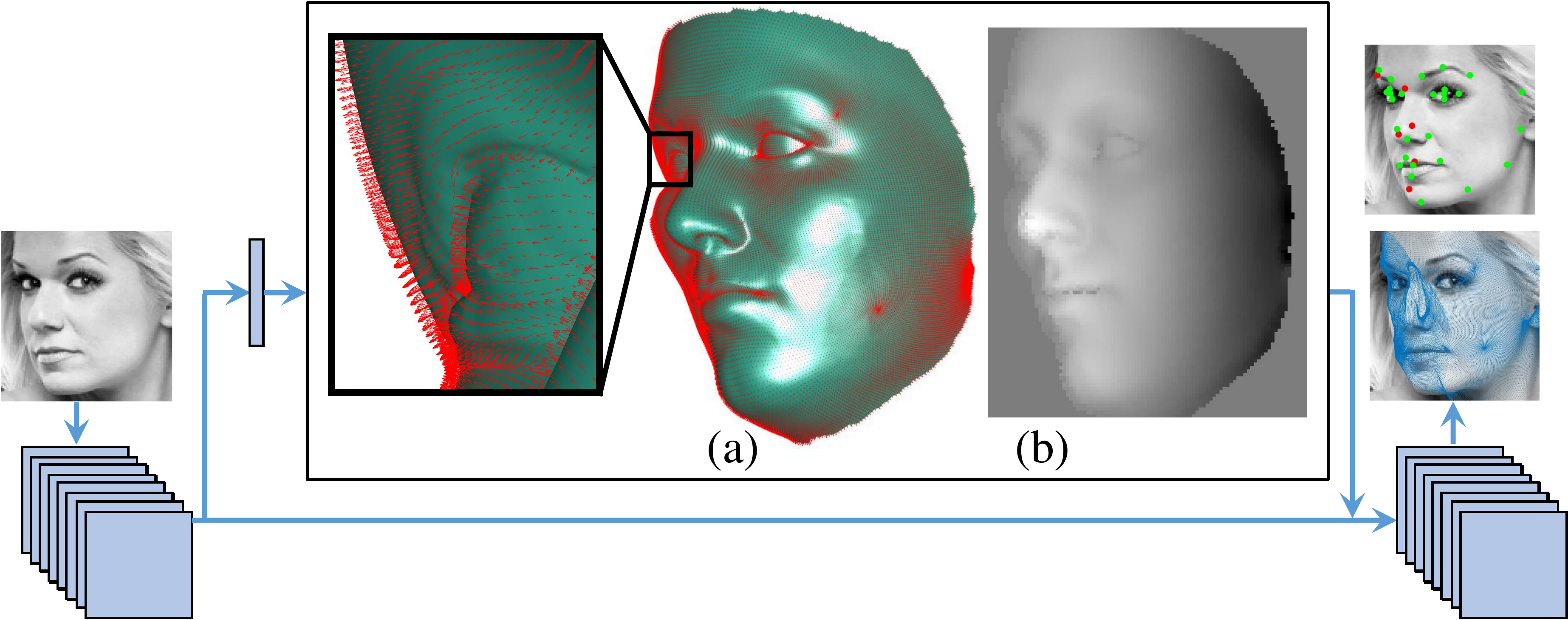}} 
\end{tabular}
\end{center}
\figvspace
   \caption{For the purpose of learning an end-to-end  face alignment model, our novel visualization layer reconstructs the $3$D face shape (a) from the estimated parameters inside the CNN and synthesizes a $2$D image (b) via the surface normal vectors of visible vertexes.} 
\label{fig:First_Image}\figvspace
\end{figure}

Despite the recent success, the cascade of CNNs, when applied to LPFA, suffers from the following drawbacks.

\textbf{Lack of end-to-end training}: 
It is a consensus that end-to-end training is desired for CNN~\cite{jaderberg2015spatial,caesar2016region}. 
However, one CNN regressor is trained independently at each cascade stage. 
Sometimes even multiple CNNs are applied independently at each stage.
E.g., locations of different landmark sets are estimated by various CNNs and combined by a separate fusing module~\cite{sun2013deep}. 
Therefore, these CNNs can not be optimized jointly and might lead to a sub-optimal solution. 

\textbf{Hand-crafted feature extraction}: 
Since the CNNs are trained independently, feature extraction is required to utilize the result of previous CNN and provide input to the current CNN. Simple feature extraction methods are used, e.g., extracting patches~\cite{zhang2014coarse,sun2013deep} based on $2$D or $3$D face shapes without considering other factors including pose and expression. 
Normally, the cascade of CNNs is a collection of shallow CNNs where each one has less than five layers. 
Hence, this framework can not extract {\it deep} features by building upon the extracted features of early-stage CNNs.

\textbf{Slow training speed}: 
Training a cascade of CNNs is usually time-consuming for two reasons. 
Firstly, the CNNs are trained sequentially, one after another. 
Secondly, feature extraction is required between two consecutive CNNs. 

To address these issues, as shown in Fig.~\ref{fig:First_Image}, we introduce a novel layer, named the visualization layer, into a CNN architecture, for the LPFA problem. 
Our CNN architecture consists of several blocks, which are called visualization blocks. 
This architecture can be considered as a cascade of shallow CNNs. 
The new layer visualizes the alignment result of the previous visualization block and utilizes it in the current block.
It is designed based on several guidelines. 
Firstly, it is derived from the surface normals of the underlying $3$D face model and encodes the relative pose between the face and camera. 
The use of surface normals is partially inspired by the success of adopting surface normals for $3$D face recognition~\cite{mohammadzade2013iterative}. 
Secondly, the visualization layer is differentiable, which allows the gradient to be computed analytically, enabling end-to-end training. 
Lastly, a mask is utilized to differentiate between pixels in the middle and contour parts of a face, and to also make the pixel values of the visualized images similar across various poses. 

Benefiting from the design of the visualization layer, our method has the following advantages and contributions:

$\diamond$  The proposed method allows a block in the CNN to utilize the extracted features from previous blocks and extract deeper features. Therefore, extraction of hand-crafted features is no longer necessary.

$\diamond$ The visualization layer is differentiable, allowing for backpropagation of an error from a later block to an earlier one. To the best of our knowledge, this is the first method for large-pose face alignment, that utilizes only one single CNN and allows end-to-end training.

$\diamond$ The proposed method converges faster during the training phase compared to the cascade of CNNs. Therefore, the training time is dramatically reduced.

The source code of the proposed method with the trained model are released at \href{http://cvlab.cse.msu.edu/project-pifa.html}{here}.



%% file: CVPR17_Bosch_prior.tex
\Section{Prior Work}
\label{Sec:PriorWork}

This section reviews the relevant prior work in three topics: cascade of regressors for face alignment, convolutional recurrent neural network and visualization in deep learning.

\Paragraph{Cascade of Regressors for Face Alignment} 
Cascade of Regressors is a classic approach in not only conventional face alignment~\cite{yu2013pose,zhu2012face}, but also the large-pose face alignment~\cite{jourabloo2015pose, hsu2015regressive,wu2015robust,zhuunconstrained}.
To handle large poses, many approaches go beyond $2$D landmarks and also estimate $3$D landmarks and $3$D face shapes~\cite{jourabloo2016large,zhu2015face}.  
Zhu et al.~\cite{zhuunconstrained} use a set of local regressors to estimate the $2$D shape update, and fuse their results with another regressor.  The occlusion-invariant approach of RCPR~\cite{burgos2013robust} is applicable to large poses since self-occlusion is one type of occlusions.
An iterative probabilistic method is utilized in~\cite{gu20063d,jeni2015dense} for registering $3$D shape to the pre-computed $2$D landmarks. 
Tulyakov et al.~\cite{tulyakov2015regressing} also use a cascade of regressors to estimate $3$D landmark updates directly from a single image. 
Some even use two regressors at each cascade stage. 
Wu et al.~\cite{wu2015robust} use one regressor to estimate the $2$D shape update and the other to estimate the visibility of each landmark. 
Similarly, Liu et~al.~\cite{liu2016joint} employ one regressor for $2$D shape update and the other uses the $2$D shape to estimate the $3$D face shape.

Among methods with cascade of regressors, CNN is a popular choice of regressors due to its strong learning ability.
These methods typically extract hand-crafted features between consecutive regressors. 
TCDCN~\cite{zhang2014facial} use one CNN to estimate five landmarks, with yaw angles within $\pm60^{\circ}$. 
A cascade of stacked autoencoder (SAE) progressively estimates $2$D landmark updates from extracted patches~\cite{zhang2014coarse}. 
Similarly, cascades of CNNs with global or local patches are combined at each stage, and their results are fused via averaging~\cite{sun2013deep,zhou2013extensive}.
The methods in~\cite{jourabloo2016large,zhu2015face} combine cascade of CNNs with $3$D feature extraction to estimate the dense $3$D face shape. All aforementioned methods lack the ability to end-to-end train the network, which is our novel contribution to large-pose face alignment.

\begin{figure*}[t]
\begin{center}
\includegraphics[width=0.85\textwidth]{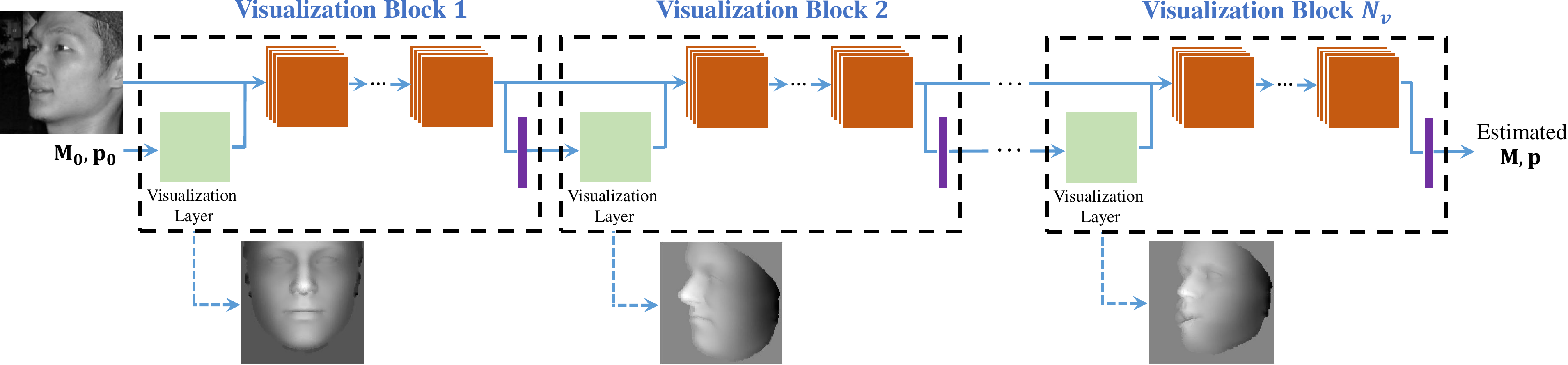}
\end{center}
\figvspace
   \caption{The proposed CNN architecture. We use green, orange, and purple to represent the visualization layer, convolutional layer, and fully connected layer, respectively. Please refer to Fig.~\ref{fig:visualization_Block} for the details of the visualization block.}
   \label{fig:CNN_overallArchitecture}\figvspace \vspace{-3mm}
\end{figure*}

\Paragraph{Convolutional Recurrent Neural Network (CRNN)} 
The face alignment methods based on the CRNNs~\cite{trigeorgis2016mnemonic,xiao2016robust,wang2016recurrent} are the first attempts to combine cascade of regressors with joint optimization, for aligning mostly frontal faces. 
Their convolutional part extracts features from the whole image~\cite{xiao2016robust} or from the patches at the landmark locations~\cite{trigeorgis2016mnemonic}. 
The recurrent part facilitates the joint optimization by sharing information among all regressors. 
The main differences between the proposed method and CRNNs are: 1) existing CRNN methods are designed for near-frontal face alignment, while ours is for LPFA; 2) the CRNN methods share the same CNN at all stages, while our CNN of each block is different which might be more suitable for estimating the coarse-to-fine mappings during the course of alignment; 3) due to our new differentiable visualization layer, our method has one additional flow of the gradient back-propagation (note the two blue arrows between consectutive blocks in Fig.~\ref{fig:CNN_overallArchitecture}).

\Paragraph{Visualization in Deep Learning}
Visualization techniques have been used in deep learning to assist in making a relative comparison among the input data and focusing on the region of interest. 
These methods can be categorized in two groups. 
The first exploits the deconvolutional and upsampling layers to either expand response maps~\cite{peng2016recurrent,li2016face} or represent estimated parameters~\cite{yang2015weakly}. 
Alternatively, various types of feature maps, e.g., heatmaps and Z-Buffering, can represent the current estimation of landmarks and parameters. 
In~\cite{wu2016single,newell2016stacked,bulat2016convolutional}, $2$D landmark heatmaps represent the landmarks' locations. \cite{bulat2016convolutional} proposes a two step large pose alignment based on heatmaps to make more precise estimations.
The heatmaps suffer from three drawbacks: 1) lack of the capability to represent objects in details; 2) requirement of one heatmap per landmark due to its weak representation power. 3) they cannot estimate visibility of landmarks. 
The Z-Buffer rendered using the estimated $3$D face is fed to the CNNs~\cite{zhu2015face} to convey the results of a previous CNN to the next one. 
However, the Z-Buffer representation is not differentiable, and hence does not allow end-to-end training. 
In contrast, our visualization layer is differentiable and encodes the face geometry details via surface normals. 
It guides the CNN to focus on the face area that incorporates both the pose and expression information. 


%% file: CVPR17_Bosch_proposed.tex
\vspace{-2mm}
\Section{Proposed Method}
\label{Sec:Method}
\vspace{-3mm}
Given a single face image with an arbitrary pose, our goal is to estimate the $2$D landmarks with their visibility labels by fitting a $3$D face model. Towards this end, we propose a CNN architecture with end-to-end training for model fitting, as shown in Fig.~\ref{fig:CNN_overallArchitecture}. 
In this section, we will first describe the underlying $3$D face model used in this work, followed by our CNN architecture and the visualization layer.
\vspace{-6mm}
\subsection{$3$D and $2$D Face Shapes}
\vspace{-2mm}
We use the $3$D Morphable Model ($3$DMM) for representing the $3$D shape of a face. 
$3$DMM represents a $3$D face $\textbf{S}_p$ as a linear combination of mean shape $\textbf{S}_0$, identity bases $\textbf{S}^I$ and expression bases $\textbf{S}^E$ as follows:
\begin{equation}
\textbf{S}_p=\textbf{S}_0+\sum_k^{N_I} p_k^I \textbf{S}_k^I+\sum_k^{N_E} p_k^E \textbf{S}_k^E.
\end{equation}
We use vector $\textbf{p}=[\textbf{p}^I, \textbf{p}^E]$ to indicate the $3$D shape parameters, where $\textbf{p}^I = [p_0^I,\cdots,p_{N_I}^I]$ are the identity parameters and  $\textbf{p}^E = [p_0^E,\cdots,p_{N_E}^E]$ are the expression parameters. We use the Basel $3$D face model~\cite{paysan20093d}, which has $199$ bases, as our identity bases and the face wearhouse model~\cite{cao2014facewarehouse} with $29$ bases as our expression bases. Each $3$D face shape consists of a set of $Q$ $3$D vertexes:
\begin{equation}
\textbf{S}_p=\left(\begin{tabular}{cccc}
  $x_1^p$ & $x_2^p$ & $\ldots$ & $x_Q^p$\\
  $y_1^p$ & $y_2^p$ & $\ldots$ & $y_Q^p$\\
  $z_1^p$ & $z_2^p$ & $\ldots$ & $z_Q^p$
\end{tabular}\right ).
\end{equation}

The $2$D face shapes are the projection of $3$D shapes. In this work, we use the weak perspective projection model with $6$ degrees of freedoms, i.e., one for scale, three for rotation angles and two for translations, which projects the $3$D face shape $\textbf{S}_p$ onto $2$D images to obtain the $2$D shape $\textbf{U}$: 
\begin{equation}
\textbf{U}=f(\textbf{P})=\textbf{M}\left(\begin{tabular}{c}
  $\textbf{S}_p(:,\textbf{b})$\\
  $\mathbf{1}$
\end{tabular}\right ),
\label{Equ:2DShape}
\end{equation}
where
\begin{equation}
\textbf{M}=\left[\begin{tabular}{cccc}
  $m_1$ & $m_2$ & $m_3$ & $m_4$\\
  $m_5$ & $m_6$ & $m_7$ & $m_8$
\end{tabular}\right],
\label{Equ:m}
\end{equation}
and
\begin{equation}
\textbf{U}=\left(\begin{tabular}{cccc}
  $x_1^t$ & $x_2^t$ & $\ldots$ & $x_N^t$\\
  $y_1^t$ & $y_2^t$ & $\ldots$ & $y_N^t$
\end{tabular}\right ).
\end{equation}    
Here $\textbf{U}$ collects a set of $N$ $2$D landmarks, $\textbf{M}$ is the camera projection matrix, with misuse of notation $\textbf{P} = \{\textbf{M}, \textbf{p}\}$, and the $N$-dim vector $\textbf{b}$ includes $3$D vertex indexes which are semantically corresponding to $2$D landmarks. 
We denote $\textbf{m}_1 = [m_1\;m_2\;m_3]$ and $\textbf{m}_2 = [m_5\;m_6\;m_7]$ as the first two rows of the scaled rotation component, while $m_{4}$ and $m_{8}$ are the translations.

Eqn.~\ref{Equ:2DShape} establishs the relationship, or equivalency, between $2$D landmarks $\textbf{U}$ and $\textbf{P}$, i.e., $3$D shape parameters $\textbf{p}$ and the camera projection matrix $\textbf{M}$. 
Given that almost all the training images for face alignment have only $2$D labels, i.e., $\textbf{U}$, we preform a data augmentation step similar to~\cite{jourabloo2016large} to compute their corresponding $\textbf{P}$.
Given an input image, our goal is to estimate the parameter $\textbf{P}$, based on which the $2$D landmarks and their visibilities can be naturally derived.

\vspace{-3mm}
\subsection{Proposed CNN Architecture}
\vspace{-2mm}

Our CNN architecture resembles the cascade of CNNs, while each ``shallow CNN" is defined as a visualization block. 
Inside each block, a visualization layer based on the latest parameter estimation serves as a bridge between consecutive blocks. 
This design enables us to address the drawbacks of typical cascade of regressors in Sec.~\ref{Sec:Intro}. 
We now describe the visualization block and CNN architecture, and dive into the details of the visualization layer in Sec.~\ref{section:Visualization}. 

\begin{figure}[t!]
\begin{center}
\includegraphics[width=0.95\linewidth]{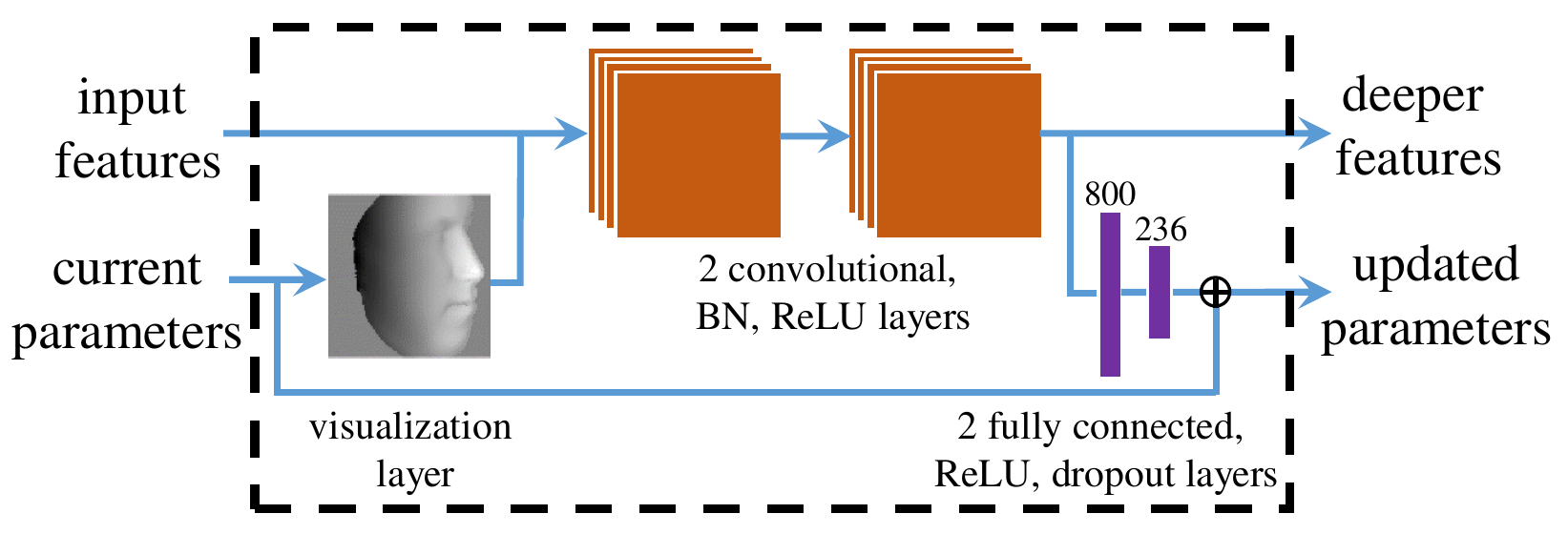}
\end{center}
\figvspace
   \caption{A visualization block consists of a visualization layer, two convolutional layers and two fully connected layers.}
\label{fig:visualization_Block}\figvspace
\end{figure}

\textbf{Visualization Block} 
Fig.~\ref{fig:visualization_Block} shows the structure of our visualization block.
The visualization layer generates a feature map based on the current estimated, or input, parameter $\textbf{P}$, and will be described in Sect.~\ref{section:Visualization}. 
Each convolutional layer is followed by a batch normalization (BN) layer and a ReLU layer, it extracts deeper features based on the input features provided by the previous visualization block and visualization layer output. 
Between the two fully connected layers, the first one is followed by a ReLU layer and a dropout layer,  while the second one simultaneously estimates the update of $\textbf{M}$ and $\textbf{p}$, $\Delta\textbf{P}$. 
The outputs of the visualization block are deeper features and the new estimation of the parameters, when adding $\Delta\textbf{P}$ to the input $\textbf{P}$.
As in Fig.~\ref{fig:visualization_Block}, basically the top part of the visualization block focuses on learning deeper features, while the bottom part utilizes such features to estimate the parameters in a ResNet-like structure~\cite{he2015deep}.
During the backward pass of the training phase, the visualization block backpropagates the loss through both of its inputs to adjust the convolutional and fully connected layers in the previous blocks.
This allows the block to extract better features that are suitable for the next block and improve the overall parameter estimation.

\textbf{CNN Architecture} 
The proposed CNN architecture consists of several connected visualization blocks as shown in Fig.~\ref{fig:CNN_overallArchitecture}. 
The inputs include the image and an initial estimation of the parameter $\textbf{P}^0$; and the output is the final estimation of the parameters. 
Compared to the typical cascade of CNNs, due to the joint optimization of all visualization blocks with backpropagation of the loss functions, the proposed architecture is able to converge in substantially fewer epochs during training. 

\textbf{Loss Functions} Two types of loss functions are employed in our CNN architecture. 
The first one is an Euclidean loss between the estimation and the target of the parameter update, with each parameter weighted separately: 
\begin{equation}
\textbf{E}_{P}^i=(\Delta\textbf{P}^i-\Delta\bar{\textbf{P}}^i)^{T}\mathbf{W}(\Delta\textbf{P}^i-\Delta\bar{\textbf{P}}^i), 
\label{Eq:EuclcideanLossOnParameter}
\end{equation}
where $\textbf{E}_{P}^i$ is the loss, $\Delta\textbf{P}^i$ is the estimation and $\Delta\bar{\textbf{P}}^i$ is the target (or ground truth) at the $i$-th visualization block. 
The diagonal matrix $\mathbf{W}$ contains the weights. 
For each element of the shape parameter $\textbf{p}$, its weight is the inverse of the standard deviation that was obtained from the data used in $3$DMM training. 
To compensate the relative scale among the parameters of $\textbf{M}$, we compute the ratio $r$ between the average of scaled rotation parameters and average of translation parameters in the training data. 
We set the weights of the scaled rotation parameters of $\textbf{M}$ to $\frac{1}{r}$ and the weights of the translation of $\textbf{M}$ to $1$.
The second type of loss function is the Euclidean loss on the resultant $2$D landmarks:
\begin{equation}
\textbf{E}_{S}^i=\|f(\textbf{P}^i+\Delta\textbf{P}^i)-\bar{\textbf{U}}\|^2,
\label{Eq:EuclcideanLossOnLandmark}
\end{equation}
where $\bar{\textbf{U}}$ is the ground truth $2$D landmarks, and $\textbf{P}^i$ is the input parameter to the $i$-th block, i.e., the output of the $i-1$-th block. 
$f(\cdotp)$ computes $2$D landmark locations using the currently updated parameters via Eqn.~\ref{Equ:2DShape}. 
For backpropagation of this loss function to the parameter $\Delta\textbf{P}$, we use the chain rule to compute the gradient (see supplemental material for the detailed derivation).
\vspace{-1mm}
\begin{equation*}
\frac{\partial \textbf{E}_{S}^i}{\partial \Delta\textbf{P}^i}=\frac{\partial \textbf{E}_{S}^i}{\partial f}\frac{\partial f}{\partial \textbf{P}^i}.
\vspace{-1mm}
\end{equation*}

For the first three visualization blocks, the Euclidean loss on the parameter updates (Eqn.~\ref{Eq:EuclcideanLossOnParameter}) is used, while the Euclidean loss on $2$D landmarks (Eqn.~\ref{Eq:EuclcideanLossOnLandmark})  is applied to the last three blocks. 
The first three blocks estimate parameters to align $3$D shape to the face image roughly and the last three blocks leverage the good initialization to estimate the parameters and the $2$D landmark locations more precisely.


\vspace{-2mm}

\subsection{Visualization Layer}
\label{section:Visualization}
\vspace{-2mm}
Several visualization techniques have been explored for facial analysis. 
In particular, Z-Buffering, which is widely used in prior works~\cite{blanz2003face,blanz1999morphable}, is a simple and fast $2$D representation for the $3$D shape. 
However, this representation is not differentiable. 
In contrast, our visualization is based on surface normals of the $3$D face, which describes surface's orientation in a local neighbourhoods. 
It has been successfully utilized for different facial analysis tasks, e.g., $3$D face reconstruction~\cite{roth2016adaptive} and $3$D face recognition~\cite{mohammadzade2013iterative}. 

In this work, we use the $z$ coordinate of surface normals of each vertex, transformed with the pose. 
It is an indicator of ``frontability" of a vertex, i.e., the amount that the surface normal is pointing towards the camera. 
This quantity is used to assign an intensity value at its projected $2$D location to construct the visualization image. 
The frontability measure $\textbf{g}$, a $Q$-dim vector, can be computated as,
\vspace{-1mm}
\begin{equation}
\textbf{g}=\max \left (\textbf{0},\frac{(\textbf{m}_1\times \textbf{m}_2)}{\|\textbf{m}_1\|\|\textbf{m}_2\|} {\textbf{N}}_0 \right ), 
\label{eqn:G} \vspace{-1mm}
\end{equation}
where $\times$ is the cross product, and $\|.\|$ denotes the $L_{2}$ norm.
The $3\times Q$ matrix ${\textbf{N}}_0$ is the surface normal vectors of a $3$D face shape.
To avoid the high computational cost of computing the surface normals after each shape update, we approximate ${\textbf{N}}_0$ as the  surface normals of the mean $3$D face.  
Note that both the face shape and pose are still continuously updated across various visualization blocks, and are used to determine the projected $2$D location. 
Hence, this approximation would only slightly affect the intensity value. 
To transform the surface normal based on the pose, we apply the estimation of the scaled rotation matrix ($\textbf{m}_1$ and $\textbf{m}_2$) to the surface normals computed from the mean face. 
The value is then truncated with the lower bound of $0$ (Eqn.~\ref{eqn:G}).

\begin{figure}
\begin{center}
\vspace{-1mm}
\includegraphics[width=0.6\linewidth]{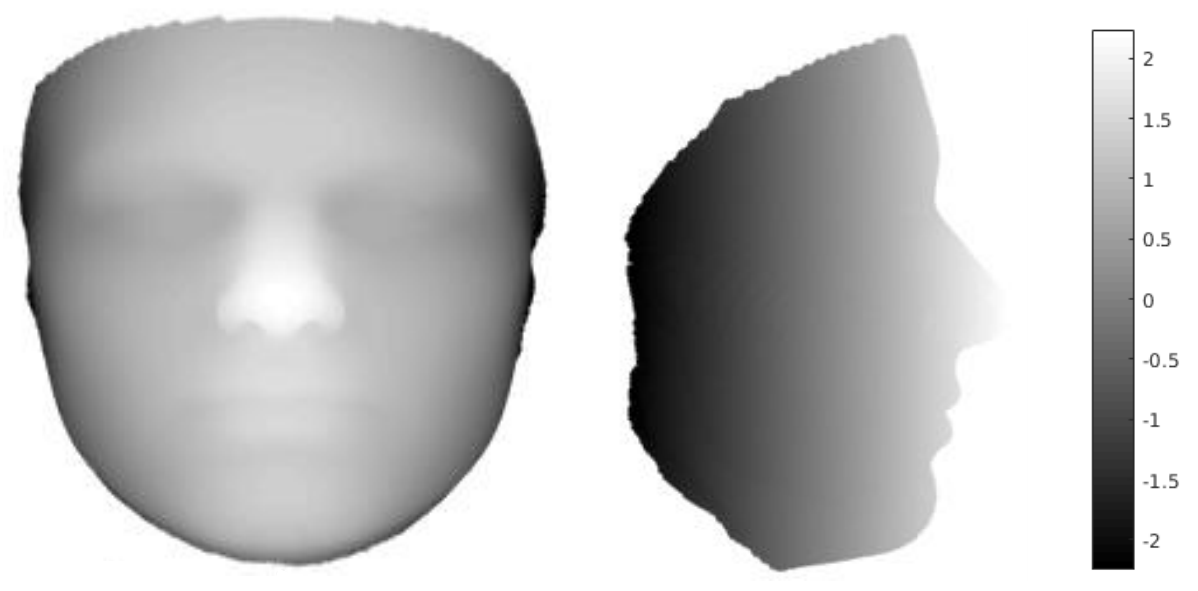} 

\end{center}
\figvspace
\caption{The frontal and side views of the mask $\textbf{a}$ that has positive values in the middle and negative values in the contour area. } 
\label{fig:Mask} \figvspace
\end{figure}

The pixel intensity of a visualized image  $\textbf{V}(u,v)$ is computed as the weighted average of the frontability measures within a local neighbourhood:
\begin{equation}
\textbf{V}(u,v)=\frac{\sum_{ q\in\mathbb{D}(u,v)} \textbf{g}(q) \textbf{a}(q) w(u,v,x_{q}^t,y_{q}^t)}{\sum_{ q\in\mathbb{D}(u,v)} w(u,v,x_{q}^t,y_{q}^t)},
\label{eqn:visalization}
\end{equation}
where $\mathbb{D}(u,v)$ is the set of indexes of vertexes whose $2$D projected locations are within the local neighborhood of the pixel $(u, v)$. 
$(x_{q}^t,y_{q}^t)$ is the $2$D projected location of $q$-th $3$D vertex.
The weight $w$ is the distance metric between the pixel $(u, v)$ and the projected location $(x_{q}^t,y_{q}^t)$, 
\begin{equation}
w(u,v,x_q^t,y_q^t)=\exp \left( -\frac{(u-x_q^t)^2+(v-y_q^t)^2}{2\sigma^2}\right).
\end{equation}
$\textbf{a}$ is a $Q$-dim mask vector with positive values for vertexes in the middle area of the face and negative values for vertexes around the contour area of the face:
\begin{equation}
\textbf{a}(q)=\exp \left( -\frac{(x^n-x_q^p)^2+(y^n-y_q^p)^2+(z^n-z_q^p)^2}{2\sigma_n^2}\right),
\end{equation}
where $(x^n,y^n,z^n)$ is the vertex coordinate of the nose tip. 
$\textbf{a}$ is pre-computed and normalized for zero-mean and unit standard deviation. 
The mask is utilized to discriminate between the central and boundary areas of the face, as well as to increase similarity across visualization of different faces. 
A visualization of the mask is provided in Fig.~\ref{fig:Mask}.

\begin{figure}[t!]
\begin{center}
\includegraphics[width=0.7\linewidth]{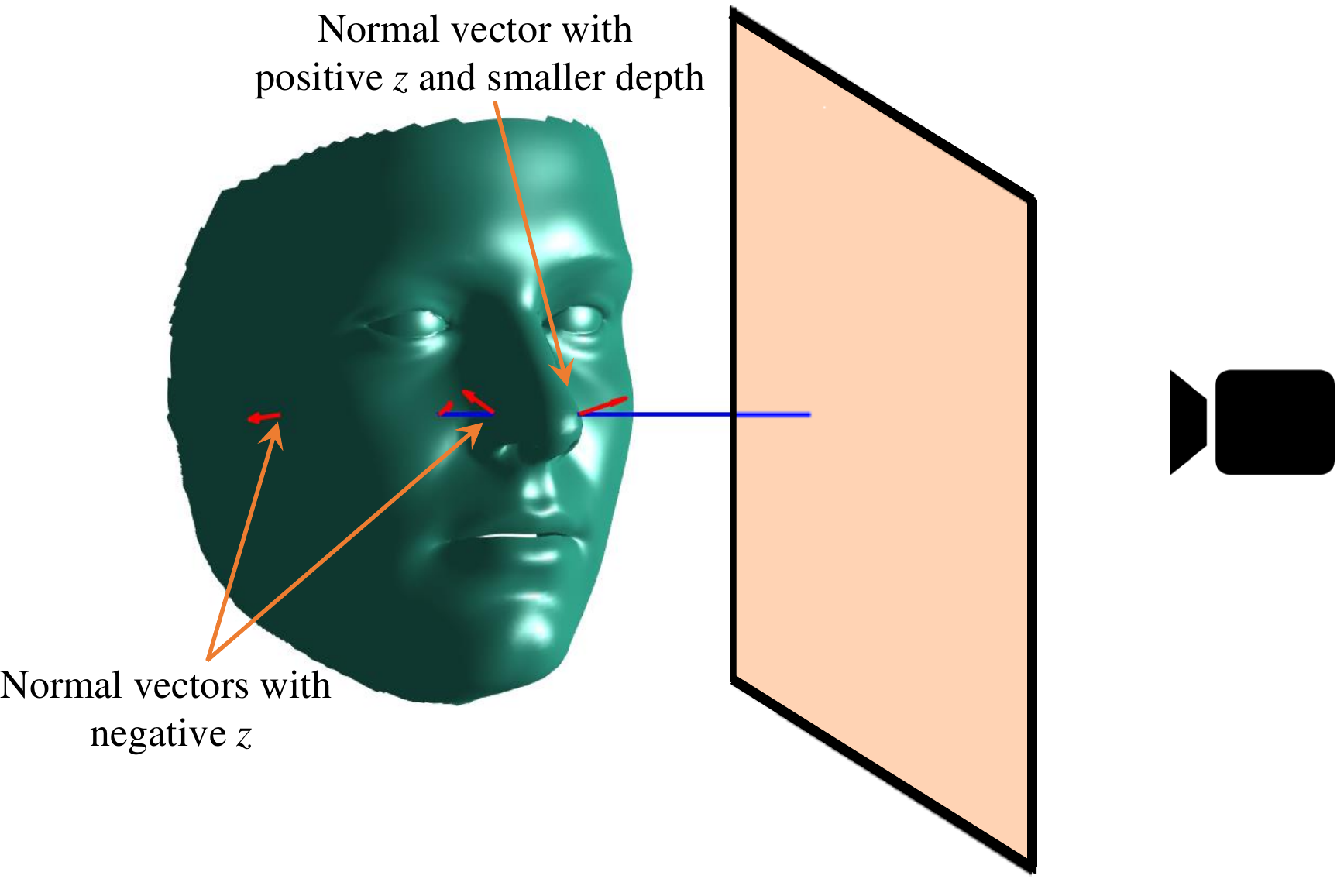}
\end{center}\figvspace
   \caption{The projections of four vertexes fall in the same image pixel. The  surface normal vectors (red arrows) of two vertexes have positive $z$ coordinates and the other two have negative $z$. Between the two vertexes with positive $z$, the one with the smaller depth (closer to the image plane) is used to fill the pixel.}
\label{fig:ZBuffering} \figvspace 
\end{figure}

Since the human face is a $3$D object, visualizing it at an arbitrary view angle requires the estimation of the visibility of each $3$D vertex.
To avoid the computationally expensive visibility test via rendering, we adopt two strategies for approximation. 
Firstly, we prune the vertexes whose frontability measures $\textbf{g}$ equal $0$, i.e., the vertexes pointing against the camera. 
Secondly, if multiple vertexes projects to a same image pixel, we keep only the one with the smallest depth values. 
An example is illustrated in Fig.~\ref{fig:ZBuffering}.  


\textbf{Backpropagation} To allow backpropagation of the loss functions through the visualization layer, we compute the derivative of \textbf{V} with respect to the elements of the parameters \textbf{M} and \textbf{p}. 
Firstly, we compute the partial derivatives, 
$\frac{\partial\textbf{g}}{\partial m_k}$, $\frac{\partial w(u,v,x_i^t,y_i^t)}{\partial m_k}$ and $\frac{\partial w(u,v,x_i^t,y_i^t)}{\partial p_j}$, then the derivatives of $\frac{\partial\textbf{V}}{\partial m_k}$ and $\frac{\partial\textbf{V}}{\partial p_j}$ can be computed based on Eqn.~\ref{eqn:visalization} (the details are provided in the supplemental material).

%% file: CVPR17_Bosch_exp.tex
\vspace{-2mm}
\Section{Experimental Results}
\label{Sec:Exp}
\vspace{-3mm}
We evaluate our proposed method on two challenging LPFA datasets, namely AFLW and AFW, both qualitatively and quantitatively, as well as the near-frontal face dataset of $300$W. 
Further, we conduct experiments on different CNN architectures to validate our visualization layer design. 

\Paragraph{Implementation details}
Our implementation is built upon the Caffe toolbox~\cite{jia2014caffe}. 
In all of the experiments, we use six visualization blocks ($N_v$) with two convolutional layers ($N_c$) and fully connected layers in each block (Fig.~\ref{fig:visualization_Block}). 
Details of the network structure are provided in Tab.~\ref{table:VBdetail}. 

Instead of using the sequentially pretrain strategy~\cite{yan2015hd}, we perform the joint end-to-end training from scratch. 
To better estimate the parameter update in each block and to increase the effectiveness of using visualization block, we set the weight of the loss function in the first visualization block to $1$, and linearly increase the weights by one for each block, i.e., the loss weight of the last block is $6$.
This strategy helps the CNN to pay more attention to the landmark loss used in later blocks. 
On the one hand, backpropagation of loss functions in the last blocks has more impact in the first block, and on the other hand the last block can adopt itself more quickly to the changes in the first block.  

In the training phase, we set the weight decay to $0.005$, the momentum to $0.99$, the initial learning rate to $1\mathrm{e}{-6}$. 
Besides, we decrease the learning rate to $5\mathrm{e}{-6}$ and $1\mathrm{e}{-7}$ after $20$ and $29$ epochs. 
In total, the training phase is continued for $33$ epochs for all experiments. 

\begin{table}[t!]\small
\caption{\small Number and size of convolutional filters in each visualization block. For all blocks, the two fully connected layers have the same length of $800$ and $236$.}\figvspace\vspace{-3mm}
\begin{center}
  \resizebox{\linewidth}{!}{
\begin{tabular}{ |c|c|c|c|c|c| } 
 \hline
 Block \# & $1$ & $2$ & $3$ & $4$ & $5,6$\\\hline
Conv. & $12$ ($5$$\times$$5$) & $20$ ($3$$\times$$3$) & $28$ ($3$$\times$$3$) & $36$ ($3$$\times$$3$) & $40$ ($3$$\times$$3$)\\
layers & $16$ ($5$$\times$$5$) & $24$ ($3$$\times$$3$) & $32$ ($3$$\times$$3$) & $40$ ($3$$\times$$3$) & $40$ ($3$$\times$$3$)\\\hline


\end{tabular}
}
\end{center}
\label{table:VBdetail}\vspace{-10mm}
\end{table}
\vspace{-1mm}
\SubSection{Quantitative Evaluations on AFLW and AFW}
\vspace{-2mm}
The AFLW dataset~\cite{kostinger2011annotated} is a very challenging dataset with large-pose face images ($\pm90^{\circ}$ yaw). 
We use the subset of this dataset released by~\cite{jourabloo2016large}, which includes $3,901$ training images  and $1,299$ testing images. 
All face images in this subset are labeled with $34$ landmarks and a bounding box.
The AFW dataset~\cite{zhu2012face} contains $205$ images with $468$ faces. 
Each face image is labeled with at most $6$ landmarks with visibility labels, as well as a bounding box. 
AFW is used only for testing in our experiments. 
The bounding boxes in both datasets are used as initilization for our algorithm, as well as the baselines. 
We crop the face image inside the bounding box and normalize it to $114 \times 114$. 
Due to the memory constraint of GPUs, we have a pooling layer in the first visualization block after the first convolutional layer to decrease the size of feature maps to half, and the input to the subsequent visualization blocks is of $57 \times 57$.
To augment the training data, we generate $20$ different variations for each training image by adding noise to the location, width and height of the provided bounding boxes.

For quantitative evaluations, we use two conventional metrics. 
The first one is Mean Average Pixel Error (MAPE)~\cite{yu2013pose}, which is the average of the pixel errors for the visible landmarks. The other one is Normalized Mean Error (NME), i.e., the average of the normalized estimation error of visible landmarks.
The normalization factor is the square root of the face bounding box size~\cite{jourabloo2015pose}, instead of the eye-to-eye distance in the frontal-view face alignment.

We compare our method with several state-of-the-art methods in LPFA. 
For AFLW, we compare with LPFA~\cite{jourabloo2016large}, PIFA~\cite{jourabloo2015pose} and RCPR~\cite{burgos2013robust} with the NME metric. 
Tab.~\ref{table:AFLWRes} shows that the proposed method achieved a higher accuracy than the baseline methods. Also, CALE~\cite{bulat2016convolutional}, a heatmap-based $2$D face alignment method, reports NME of $2.96\%$ on the AFLW. We discuss about the advantage of the proposed method over heatmap-based methods in section~\ref{Sec:PriorWork}.
To demonstrate the capabilities of each visualization block, the NME computed using the estimated $\mathbf{P}$ after each block is shown in Tab.~\ref{table:AFLWStage}. 
If a higher alignment speed is desirable, it is possible to skip the last two visualization blocks with a reasonable NME.



On the AFW dataset, the comparisons are conducted with LPFA~\cite{jourabloo2016large}, PIFA~\cite{jourabloo2015pose}, CDM~\cite{yu2013pose} and TSPM~\cite{zhu2012face} with the MAPE metric. 
The evaluations are provided in Tab.~\ref{table:AFWRes}, which also shows the superiority of the proposed method.

\begin{table}[t!]\small
\caption{NME ($\%$) of four methods on AFLW dataset.}
\begin{center}
\begin{tabular}{ c|c|c|c } 
 \hline
 Proposed method & LPFA~\cite{jourabloo2016large} & PIFA & RCPR \\ 
 \hline
 $4.45$ & $4.72$ & $8.04$ & $6.26$ \\
 \hline  
\end{tabular}
\end{center}
\label{table:AFLWRes}\vspace{-7mm}
\end{table}


\begin{table}[t!]\small
\caption{NME ($\%$) of the proposed method at each visualization block on AFLW dataset. The initial NME is 25.8$\%$.}
\vspace{-3mm}
\begin{center}
  \resizebox{0.9\linewidth}{!}{
\begin{tabular}{ c|c|c|c|c|c|c } 
 \hline
 Block \# & $1$ & $2$ & $3$ & $4$ & $5$& $6$\\\hline
 NME & $9.26$ & $6.77$ & $5.51$ &$4.98$ &$4.60$ &$4.45$\\\hline
 
\end{tabular}
}
\end{center}
\label{table:AFLWStage}\vspace{-6mm}
\end{table}

\begin{figure*}[h!]\small
\begin{center}
\includegraphics[width=0.95\linewidth]{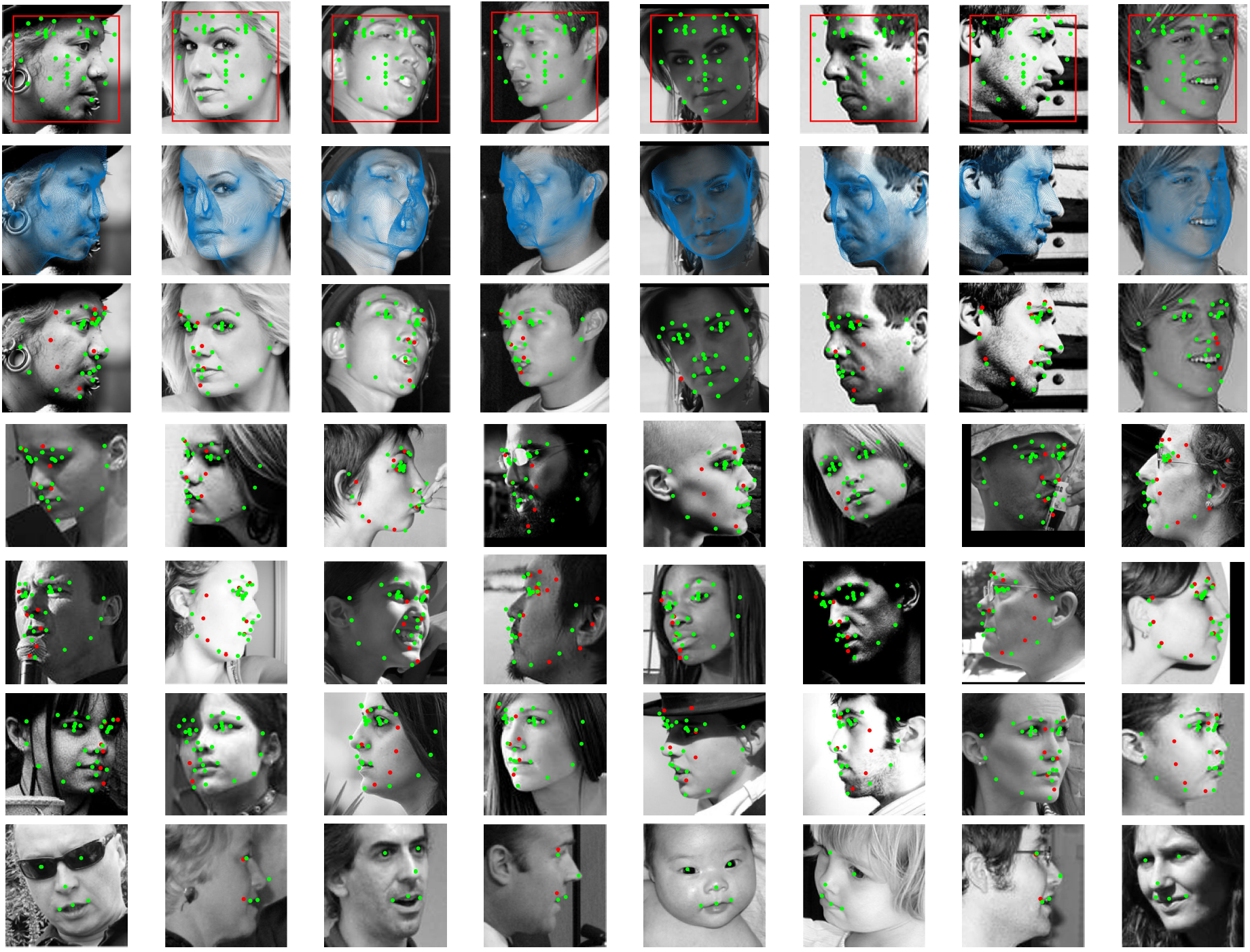} 
\end{center}

   \caption{Results of alignment on AFLW and AFW datasets, green landmarks show the estimated locations of visible landmarks and red landmarks show estimated locations of invisible landmarks. First row: provided bounding box by AFLW with initial locations of landmarks, Second: estimated $3$D dense shapes, Third: estimated landmarks, Fourth to sixth: estimated landmarks for AFLW, Seventh: estimated landmarks for AFW.}
   \label{figure:ResAFLWAFW1}
\end{figure*}
\begin{figure*}[h]\small
\begin{center}
\includegraphics[width=0.95\linewidth]{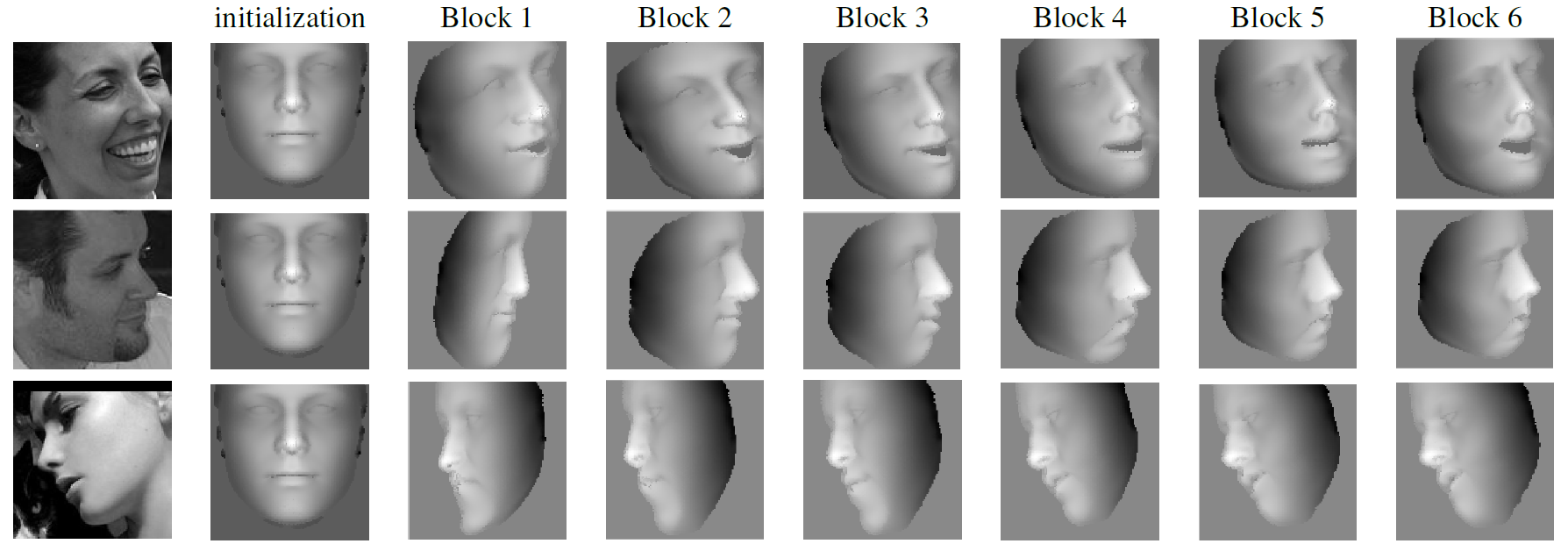} 
\end{center}

\caption{Three examples of outputs of visualization layer at each visualization block. The first row shows that the proposed method recovers the expression of the face gracefully, the third row shows the visualizations of a face with a more challenging pose.}
\label{figure:ResVisEach1}
\end{figure*}

\begin{table}[t!]\small
\caption{MAPE of five methods on AFW dataset.}
\begin{center}
\begin{tabular}{ c|c|c|c|c } 
 \hline
 Proposed method & LPFA~\cite{jourabloo2016large} & PIFA & CDM & TSPM \\ 
 \hline
 $6.27$ & $7.43$ & $8.61$ & $9.13$ & $11.09$ \\
 \hline  
\end{tabular}
\end{center}
\label{table:AFWRes}\vspace{-9mm}
\end{table}
Some examples of alignment results of the proposed method on AFLW and AFW datasets are shown in Fig.~\ref{figure:ResAFLWAFW1}. Three examples of visualization layer output at each visualization block are shown in Fig.~\ref{figure:ResVisEach1}. 

\vspace{-2mm}
\SubSection{Evaluation on 300W dataset}
\vspace{-2mm}
While our main goal is LPFA, we further evaluate on the most widely used near frontal $300$W dataset~\cite{sagonas2013300}. 
$300$W containes $3,148$ training and $689$ testing images, which are divide into common and challenging sets with $554$ and $135$ images, respectively. 
Tab.~\ref{table:300WRes} shows the NME (normalized by the interocular distance) of the proposed and state-of-the-art methods. 
The most related method to ours is $3$DDFA~\cite{zhu2015face}, which also estimates $\textbf{M}$ and $\textbf{p}$. 
Our method outperforms it on both common and challenging sets. Other near frontal alignment methods do not employ shape constraints e.g., $3$DMM which is an advantage for them. Because the span of the $3$D shape bases cannot cover all possible locations of landmarks. 
To comapre with the MDM~\cite{trigeorgis2016mnemonic}, we compute the failure rate with threshold of $0.08$. The failure rates of our method are $16.83\%$ ($6.80\%$ for MDM) and $8.99\%$ ($4.20\%$ for MDM) with $68$ and $51$ landmarks. 

\begin{table}[t!]\small
\caption{The NME of different methods on $300$W dataset.}
\begin{center}
\begin{tabular}{ c|c|c|c } 
 \hline
 Method & Common & Challenging & Full \\ 
 \hline
 ESR~\cite{cao2014face} & $5.28$ & $17.00$ & $7.58$ \\
 RCPR~\cite{burgos2013robust} & $6.18$ & $17.26$ & $8.35$ \\
 SDM~\cite{xiong2013supervised} & $5.57$ & $15.40$ & $7.50$ \\
 LBF~\cite{ren2014face} & $4.95$ & $11.98$ & $6.32$ \\
 CFSS~\cite{zhu2015face} & $4.73$ & $9.98$ & $5.76$ \\
 RCFA~\cite{wang2016recurrent} & $4.03$ & $9.85$ & $5.32$ \\
 RAR~\cite{xiao2016robust} & $4.12$ & $8.35$ & $4.94$ \\
 3DDFA~\cite{zhu2015face} & $6.15$ & $10.59$ & $7.01$ \\
 3DDFA+SDM & $5.53$ & $9.56$ & $6.31$ \\
 \hline  
 Proposed method & $5.43$ & $9.88$ & $6.30$ \\
 \hline  
\end{tabular}
\end{center}
\label{table:300WRes}\vspace{-10mm}
\end{table}

\vspace{-2mm}
\SubSection{Analysis of the Visualization Layer}
\vspace{-2mm}
We perform four sets of experiments to study the properties of the visualization layer and network architectures. 

\Paragraph{Influence of visualization layers}
To analyze the influence of the visualization layer in the testing phase, we add $5\%$ noise to the fully connected layer parameters of each visualization block, and compute the alignment error on the AFLW test set. 
The NMEs are [$4.46$, $4.53$, $4.60$, $4.66$, $4.80$, $5.16$] when each block is modified seperately. 
This analysis shows that visualized image has more influence on the later blocks, since imprecise parameters of early blocks could be compensated by later blocks. 
To evaluate the influence of the visualization layer in the training phase, we train the network without any visualization layer. 
The final NME on AFLW is $7.18\%$ which shows the importance of visualization layers for guiding the network training.

\begin{figure}[t!]\small
\begin{center}
\begin{tabular}{cc}
(a) & \includegraphics[width=0.82\linewidth]{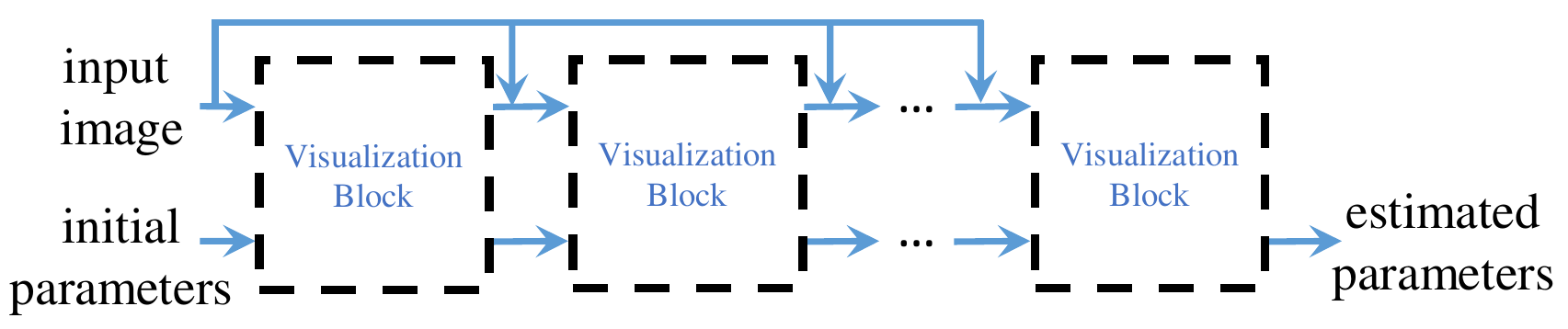} \\
(b) & \includegraphics[width=0.82\linewidth]{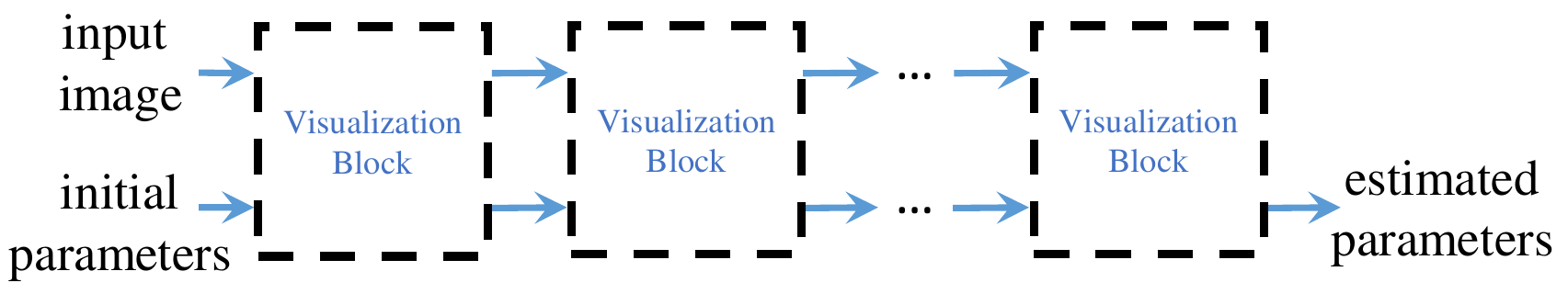} \\
(c) & \includegraphics[width=0.82\linewidth]{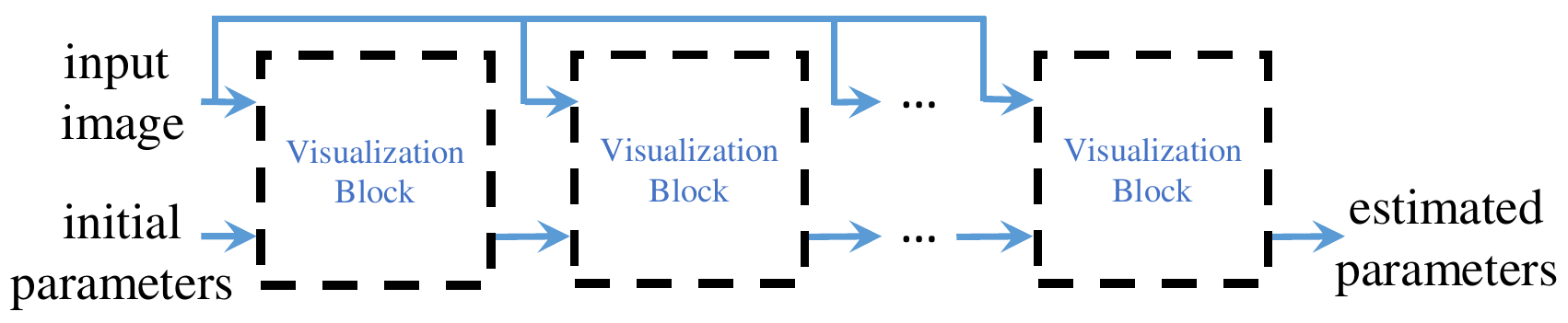}
\end{tabular}
\end{center}
\figvspace
\caption{Architectures of three CNNs with different inputs.} 
\label{fig:Arc}\figvspace
\end{figure}

\Paragraph{Advantage of deeper  features}
We train three CNN architectures (Fig.~\ref{fig:Arc}) on AFLW. 
The inputs of the visualization block in the first architecture are the input images $\textbf{I}$, feature maps $\textbf{F}$ and the visualization image $\textbf{V}$. The inputs of the second and the third architectures are $\{\textbf{F}, \textbf{V}\}$ and $\{\textbf{I},\textbf{V}\}$, respectively. 
The NME of each architecture is shown in Tab.~\ref{table:diffArch}.
While the first one performs the best, the substantial lower performance of the third one demonstrates the importance of deeper features learned across blocks.

\begin{figure}\tiny
\figvspace\vspace{-2mm}
\begin{center}
\begin{tabular}{@{}ccc@{}}
$\quad$ Input image filters & $\quad$ Visualization filters & $\qquad$ Feature maps filters\\
\includegraphics[width=0.3\linewidth]{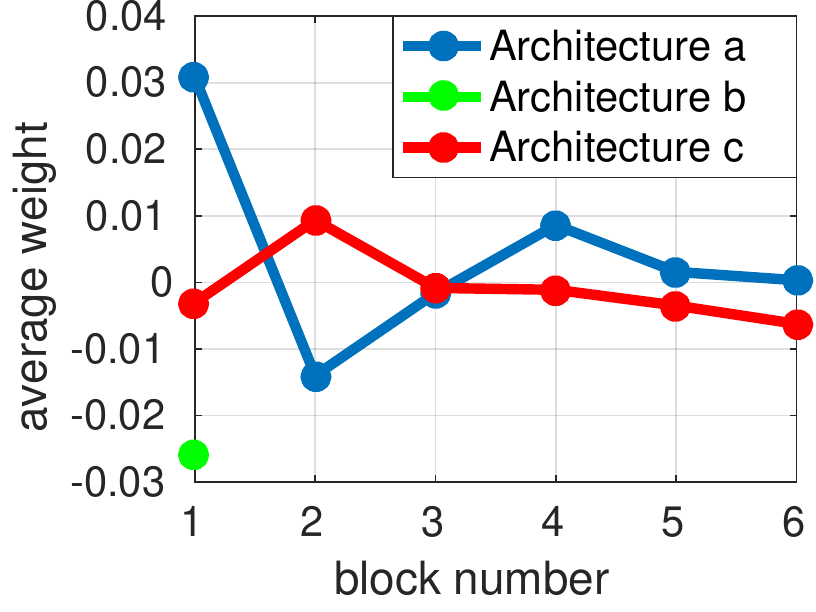} &
\includegraphics[width=0.3\linewidth]{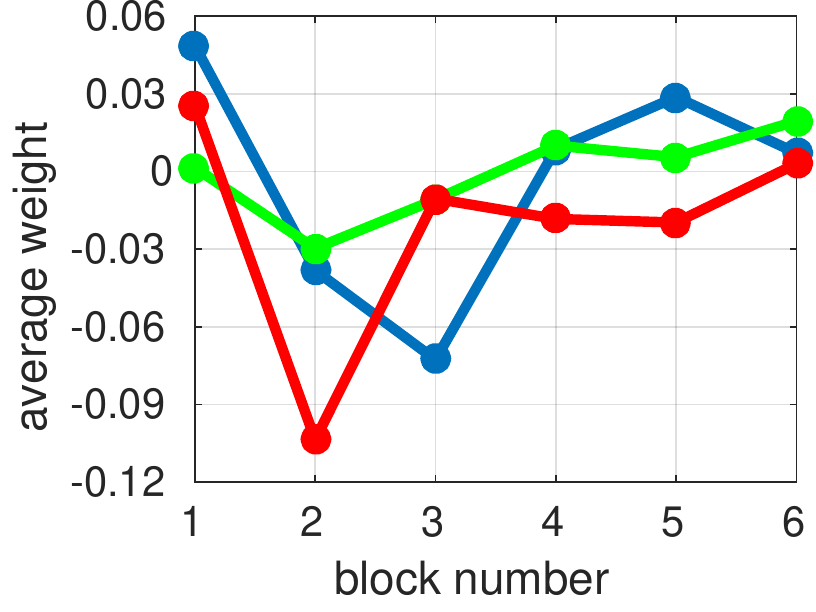} &
\includegraphics[width=0.3\linewidth]{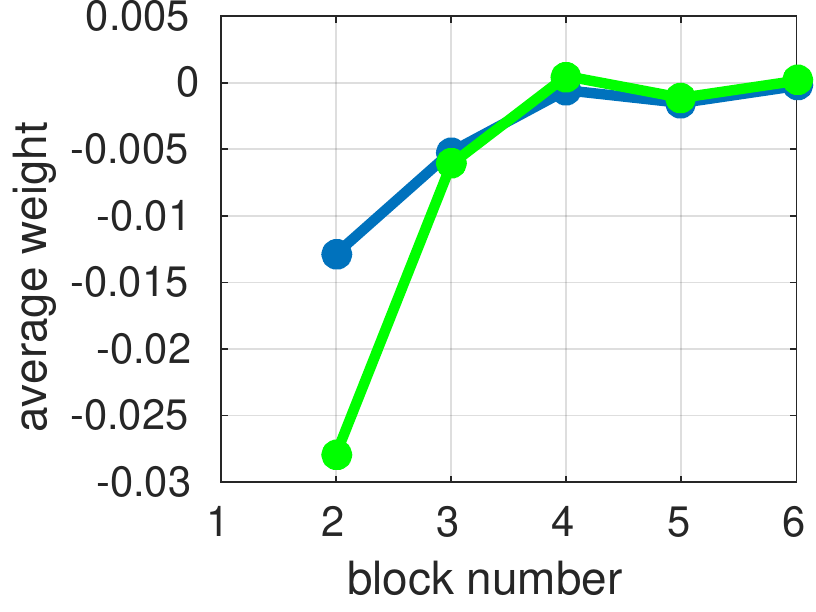}


\end{tabular}
\end{center}
\figvspace
\caption{The average of filter weights for input image, visualization and feature maps in three architectures of Fig.~\ref{fig:Arc}. The $y$-axis and $x$-axis shows the average and the block index, respectively.} 
\label{fig:Arc_Analysis}\vspace{-7mm}

\end{figure}

At the first convolutional layer of each visualization block, we compute the average of the filter weights, across both the kernel size and number of maps.
The averages for three types of input features are shown in Fig.~\ref{fig:Arc_Analysis}.
As can be observed, from the first to the sixth block, the weights continue to decrease, making a more precise estimation of small-scale parameter updates. 
Considering the number of filters in Tab.~\ref{table:VBdetail}, the total impact of feature maps are higher than the other two inputs in all blocks. 
This again shows the importance of deeper features in guiding the network to estimate parameters. 
Furthermore, the average of the visualization filter is higher than that of the input image filter, which validates the stronger influence of the proposed visualization during training. 

\begin{table}[t!]\small
\caption{The NME ($\%$) of three architectures with different inputs ($\textbf{I}$: Input image, $\textbf{V}$: Visualization, $\textbf{F}$: Feature maps).}\figvspace
\begin{center}
\begin{tabular}{ c|c|c } 
 \hline
 Architecture a & Architecture b & Architecture c \\ 
 $(\textbf{I},\textbf{F},\textbf{V})$ & $(\textbf{F}, \textbf{V})$ & $(\textbf{I},\textbf{V})$\\
 \hline
 $4.45$ & $4.48$ & $5.06$ \\
 \hline  
\end{tabular}
\end{center}
\label{table:diffArch}
\end{table}

\begin{figure}[t!]\small
\begin{center}
\includegraphics[width=0.6\linewidth]{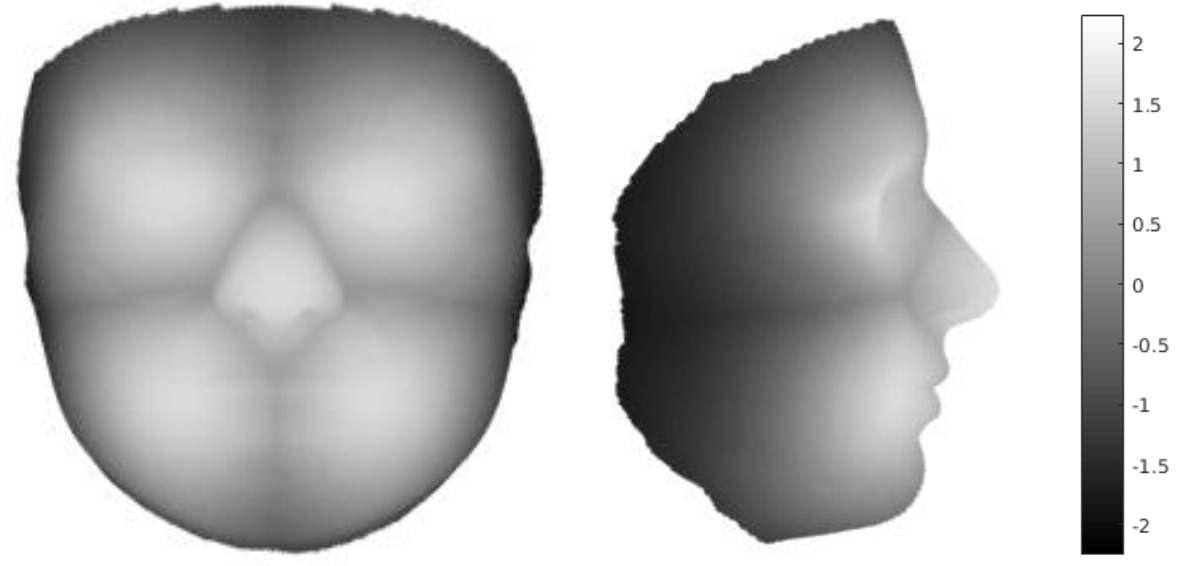} 
\end{center}
\figvspace
\caption{Mask $2$, a different designed mask with five positive areas on the eyes, top of the nose and sides of the lip.} 
\label{fig:Mask2} \vspace{-2mm}
\end{figure}

\Paragraph{Advantage of using masks}
To show the advantage of using the mask in the visualization layer, we conduct an experiment with different masks. 
Specifically, we define another mask for comparison, which is shown in Fig.~\ref{fig:Mask2}. 
It has five positive areas, i.e., the eyes, nose tip and two lip corners. 
The values are normalized to zero-mean and unit standard deviation. 
Compared to the original mask in Fig.~\ref{fig:Mask}, this mask is more complicated and conveys more information about the informative facial areas to the network. 
Moreover, to show the necessity of using the mask, we also test using visualization layers without any mask. 
The NMEs of the trained networks with different masks are shown in Tab.~\ref{table:mask}. 
Comparing the first and third columns shows the advantage of using the mask in the network. The mask makes the pixel value of visualized images to be similar for faces with different poses and discriminate between the middle-area and contour-area of the face. 
By comparing the first and second columns, we can see that utilizing more complicated mask does not further improve the result, meaning the original mask provides sufficient information for its purpose. 

\Paragraph{Different numbers of blocks and layers}
Given the total number of $12$ convolutional layers in our network, we can partition them to visualization blocks in various sizes.
To compare their performance, we train two additional CNNs, one with $4$ visualization blocks and each with $3$ convolutional layers; and the other with $3$ block and $4$ convolutional layers per block, where all three architectures have $12$ total convolutional layers.
The NME of these architectures are shown in Tab.~\ref{table:VBNum}. 
It shows the same conclusion as in~\cite{burgos2013robust} that the number of regressors is important for face alignment and we can potentially achieve a higher accuracy by increasing the number of visualization blocks. 

\begin{table}[t!]\small
\caption{NME ($\%$) of utilizing different masks.}
\begin{center}
\begin{tabular}{ c|c|c } 
 \hline
 Mask $1$ & Mask $2$ & No Mask \\ 
 \hline
 $4.45$ & $4.49$ & $5.31$ \\
 \hline  
\end{tabular}
\end{center}
\label{table:mask}\figvspace
\vspace{-3mm}
\end{table}

\begin{table}[t!]\small
\caption{NME ($\%$) of utilizing different numbers of visualization blocks ($N_v$) and convolutional layers ($N_c$).} 
\figvspace
\begin{center}
\begin{tabular}{ c|c|c } 
 \hline
 $N_v=6$ , $N_c=2$ &  $N_v=4$ , $N_c=3$ &  $N_v=3$ , $N_c=4$ \\ 
 \hline
 $4.45$ & $4.61$ & $4.83$ \\
 \hline  
\end{tabular}
\end{center}
\label{table:VBNum}\figvspace\vspace{-6mm}
\end{table}

\vspace{-2mm}
\SubSection{Time complexity} 
\vspace{-2mm}
Compared to the cascade of CNNs, one of the main advantages of end-to-end training a single CNN is the reduced training time. 
The training of the proposed method needs $33$ epochs and takes around $2.5$ days. 
The state of the art~\cite{jourabloo2016large}, that uses the same train and test sets as ours, trains six CNNs and each needs $70$ epochs. 
The total time of~\cite{jourabloo2016large} is around $7$ days. 
Similarly, the method in~\cite{zhu2015face} needs around $12$ days to train three CNNs each one with $20$ epochs, despite using different training data. 
Compared to~\cite{jourabloo2016large}, the proposed method reduces the training time by more than half.
The testing speed of proposed method is $4.3$ FPS on a Titan X GPU. It is much faster than the $0.6$ FPS speed of~\cite{jourabloo2016large} and is simalar to $4$ FPS speed of~\cite{xiao2016robust}.

\vspace{-2mm}

%% file: CVPR17_Bosch_con.tex
\Section{Conclusions}
\vspace{-2mm}
\label{Sec:Conclude}
We propose a large-pose face alignment method with end-to-end training in a single CNN. 
We present a differentiable visualization layer, which is integrated to the network and enables joint optimization by backpropagating the error from a later visualization blocks to early ones. 
It allows the visualization block to utilize the extracted features from previous blocks and extract deeper features, without extracting hand-crafted features. 
Also, the proposed method converges faster during the training phase compare to the cascade of CNNs. 
Finally, we demonstrate the superior results of the proposed method over the state-of-the-art methods.